# Prompting ChatGPT for Chinese Learning as L2:
# A CEFR and EBCL Level Study

**Miao Lin-Zucker**
CEL, University of Lyon 3, France

**Joël Bellassen**
PLIDAM, INALCO, France

**Jean-Daniel Zucker*[1]**
IRD, Sorbonne University, UMMISCO, Paris, France

**Abstract**
The use of chatbots in language learning has evolved significantly since the 1960s, becoming more sophisticated platforms as generative AI emerged. These tools now simulate natural conversations, adapting to individual learners' needs, including those studying Chinese. Our study explores how learners can use specific prompts to engage Large Language Models (LLM) as personalized chatbots, aiming to target their language level based on the *Common European Framework of Reference for Languages* (CEFR) and the *European Benchmarking Chinese Language* (EBCL) project. Focusing on A1, A1+ and A2 levels, we examine the teaching of Chinese, which presents unique challenges due to its logographic writing system. Our goal is to develop prompts that integrate oral and written skills, using high-frequency character lists and controlling oral lexical productions. These tools, powered by generative AI, aim to enhance language practice by crossing lexical and sinographic recurrence. While generative AI shows potential as a personalized tutor, further evaluation is needed to assess its effectiveness. We conducted a systematic series of experiments using ChatGPT models to evaluate their adherence to constraints specified in the prompts. The results indicate that incorporating level A1 and A1+ characters, along with the associated reference list, significantly enhances compliance with the EBCL character set. Properly prompted, LLMs can increase exposure to the target language and offer interactive exchanges to develop language skills.

---

[1] Corresponding author. Email: jean-daniel.zucker@ird.fr

# 1. Introduction

ChatGPT is arguably the most advanced chatbot today in terms of natural language understanding and generation, offering versatile assistance for various communication and learning tasks (Li et al., 2024). It is used daily by millions of people worldwide, and the question of chatbots' relevance for language teaching in general, and Chinese in particular, is central to the debate on the integration of artificial intelligence technologies in education. These tools, thanks to their adaptability, could transform the way languages are taught and learned, notably by facilitating personalized learning paths and offering immersive and interactive practice (Imran, 2023; Glazer, 2023; Huang, 2022).

Literally, a chatbot is "*a computer program designed to simulate conversations with human users, especially on the Internet*". Among the seminal works, even before the internet, was ELIZA, developed by Joseph Weizenbaum in 1966 (Weizenbaum, 1966). Although initially designed to simulate basic human conversation, this program revealed the immense potential of human-machine interactions for language practice and laid the groundwork for a new pedagogical approach. ELIZA did not truly understand conversations but identified words or phrases in the input text and responded using predefined rules that transformed or reformulated statements. Early chatbots for language learning truly emerged in the 1990s. ALICE (Artificial Linguistic Internet Computer Entity), launched in 1995 by Richard Wallace, was one of the most influential chatbots, as it introduced significant advances in conversational agent technology (Wallace, 2009). ALICE used a rule-based model (in AIML language), allowing more natural interaction with users compared to earlier models like ELIZA. ALICE also stood out for its ability to handle a vast knowledge base and produce more varied and complex responses, marking a key milestone in the development of modern chatbots. In the 2000s, early chatbots appeared on platforms like Duolingo, offering interactive language practice through scripts and decision trees, though still with limited interactions.

The real leap forward in chatbot evolution occurred between 2010 and 2020, thanks to AI. New generations of chatbots, now capable of generating more natural and adaptive responses, have transformed the user experience. Advanced algorithms enabled better context understanding and provided smoother interactions. Wang (2024) conducted a meta-analysis of 70 effect sizes from 28 studies, showing that using chatbots produced an overall positive effect on language learning performance compared to non-chatbot conditions. However, the advent of generative AI from 2020 onwards has radically changed the potential of chatbots in language learning. Even a recent comprehensive history of chatbots from 2020 (Adamopoulou, 2020) does not mention the paradigm shift brought by LLMs such as ChatGPT, which emerged after its publication.

Advanced models like GPT-3 and its successors, GPT-4, have enabled more sophisticated and personalized interactions (Kalyan, 2024). They adjust language level, communication style, and topics according to each learner's needs. These AIs simulate realistic conversations, offering language practice close to real-world situations, whether daily or professional. While generative AI promises to revolutionize language learning by offering increasingly personalized, rich, and realistic interactions, one of ChatGPT's key appeals lies in its ease of use and its vast user base. ChatGPT is already widely used by teachers in Chinese language classrooms. In (Chen et al., 2023), the authors examine how ChatGPT can be employed as a pedagogical tool to enhance Chinese language learning, particularly for conversational interactions and grammar practice. They highlight the benefits of this tool for providing instant feedback and encouraging learner engagement. In (Jiang et al., 2024), the authors focus on how ChatGPT-based activities can develop learners' critical thinking. They show that using interactive scenarios and virtual debates stimulates learners' analytical and reflective abilities. In (Li et al., 2023), the authors explore the use of ChatGPT for differentiated instruction, allowing teachers to create exercises tailored to learners' specific needs and levels. This personalized instructional model, facilitated by ChatGPT, improves learning path customization and helps manage skill differences within a class. These studies are based on the use of *prompts* (see Table 1) according to the tasks performed.

However, role instruction (see Table 1) can also be used to assign ChatGPT the role of a Chinese teacher. For example, on the Hugging Face platform, one of the assistants (see Table 1) has a simple system prompt that reads: "You are a very experienced Chinese professor teaching Chinese; you will reply to any questions in Chinese characters, Chinese pinyin, and English translation, giving also communication and learning tips. Use any information you have access to." (https://huggingface.co/chat/). Indeed, generative AI offers an unprecedented opportunity to create highly customizable virtual tutors for language learning. By allowing learners to configure the chatbot via specific system prompts, these tools can be transformed into tailored pedagogical assistants.

Table 1

**Glossary of LLM and prompt related terms**

| Term | Definition |
| --- | --- |
| *LLM* | An LLM (Large Language Model) is an artificial intelligence model based on deep neural networks, trained on massive text corpora to capture complex linguistic relationships and generate natural language based on received contexts and queries. |
| *Chatbot* | A computer program designed to simulate a conversation with human users, especially on the internet. ChatGPT is a very general example, but many others exist that are much more targeted. |
| *Foundation Model* | A foundation model is a pre-trained LLM on a vast amount of unlabeled data, capable of adapting to a variety of specific tasks through additional adjustments (fine-tuning). These models, like GPT, LLAMA, QWEN, BERT or MISTRAL are called "foundation" because they serve as a basis for developing applications in various fields. |
| *Assistant* | In the context of LLMs, an assistant is defined as an artificial intelligence model designed to respond to specific queries via a system prompt or predefined parameters. These instructions guide the model's behavior and allow it to provide a contextualized user experience. |
| *Prompt* | Prompts provide instructions to an LLM to impose rules, automate tasks, and guarantee particular qualities (and quantities) in the generated output. They also function as a form |

| | of programming, enabling the customization of both the outputs and interactions with the LLM. |
|---|---|
| *System Prompt* | A system prompt is an initial instruction given to an LLM to define its overall behavior, such as the tone or style of the responses. It sets guidelines to align the model's responses with the desired objectives during interactions. It may start with a role instruction. |
| *Prompt Engineering* | Prompt engineering is the technique of formulating specific instructions to condition and optimize the output of a language model, exploiting the internal mechanisms of the architecture to modulate its behavior and maximize the relevance of the results. |
| *Role Instruction* | Role instruction is a part of the prompt that explicitly defines the role or behavior the chatbot should adopt in a given interaction. It's an instruction that guides the model on how to act or respond based on the desired context. For example, "You are a professional Chinese language teacher, guiding the user through language learning with clear explanations and practical examples". |
| *Fine-Tuning* | Adjusting a pre-trained LLM consists of specializing it for a specific task or domain by retraining it on a small set of labeled data. This optimizes its performance for precise tasks like providing responses that satisfy character frequency constraints. |
| *Instruction Tuning* | Instruction tuning is the adjustment of a language model (LLM) so that it accurately responds to natural language instructions. This involves training the model on pairs of instructions and responses, making it more effective at providing relevant answers and following specific commands in practical applications. |
| *Instruction Deviation* | An instruction deviation occurs when a language model (LLM) does not correctly follow a given directive in a prompt, by omitting a task, executing it incorrectly, or producing results that do not conform to expectations. This can be caused by ambiguities in the prompt, limitations of the model in understanding the context, or a lack of alignment with the desired objectives. |

Large language models (LLMs) can be utilized in many ways to build a chatbot. One possible approach is to start from a *foundational model* and *fine-tune* it (see Table 1) to create a Chinese language chatbot. This approach is feasible if the models are open weight, meaning that their weights (parameters) are shared with the community, or, in the case of close-weight models, provided that one is willing to pay for the training. However, beyond the need for training data, the model must also be deployed to be usable by learners. Not all learners have access to a machine powerful enough to run these large models locally, although it is conceivable that, in the future, as model sizes decrease, this could become possible (Hutson, 2024).

The approach we are exploring is simpler; it relies on using a *system prompt* (see Table 1) that allows instructions to be given to the LLM. Our approach is also based on the recommendations of the *Common European Framework of Reference for Languages* (CEFR), which defines language skills from levels A1 to C2, with particular attention to compliance with the *European Benchmarking Chinese Language* (EBCL, 2012) project, the first European framework specifically for Chinese language competencies. In this article, we focus on CEFR levels A1, A1+, and A2, drawing on character lists established by the EBCL project, selected based on frequency and lexical combination criteria. The ambition of our research is to leverage generative AI to develop *system prompts* dedicated to Chinese language learning and acquisition, addressing the challenge of integrating written language skills in a balanced manner while adhering to the defined sinographic thresholds. We first focus on the initial threshold of high-frequency characters (320 characters for A1 level) before progressing to the second threshold (630 characters for A2 level) (Guder, 2014; Guder & Chinesisch 2015). These thresholds are established to ensure that the skills being worked on remain appropriate to the intended level: the language used in A1 level exercises must correspond to A1 vocabulary, and similarly for A2 level.

The use of character thresholds is inspired by the work of Bellassen (2009), particularly with his so-called "snowball" method, which emphasizes the idea of "recycling" previously learned characters or those belonging to a given set of frequent characters to create texts that promote character recurrence through possibly new compound words that are part of the relevant threshold. This principle is based on the pedagogical efficacy of using words composed only of characters already known to the learner, thereby reducing the cognitive load associated with acquiring new elements. This approach not only reinforces existing knowledge but also maximizes lexical combinations while minimizing memorization efforts. In this sense, this method makes learning Chinese smoother and more accessible, especially for beginners.

A chatbot like ChatGPT can be programmed through instructions — provided as a system prompt by the learner — to cover specific vocabularies, allowing the chatbot to act as a personalized tutor. We present the results of experiments based on the analysis of adherence to these instructions — *instruction deviation* (see Table 1) — by ChatGPT, for both specific language levels (A1, A1+, A2 in the CEFR) and specific activities such as reading comprehension (general, orientation reading, information & argument reading, instruction reading), creative writing, and written interaction (correspondence, notes, SMS, signs, etc.). While the contribution of this type of interaction to learning still needs to be precisely evaluated from the learner's perspective (Chen et al., 2023), a chatbot based on a large language model like ChatGPT can increase exposure to the target language and create opportunities for interactive exchanges in a foreign language, thus promoting the practice of various language skills (Wang, 2024).

The contributions of this article are:

- A reflection on the relationship between the use of generative AI for Chinese as a Foreign Language (CFL) within the CEFR framework as defined by the EBCL project and for activities related to Chinese reading and writing comprehension.
- Adapted and tested *system prompts* to meet, as much as possible, the CEFR framework for levels A1, A1+, and A2.
- A statistical analysis of *instruction deviations*, i.e., adherence to constraints regarding the characters composing the words in the chatbot's sentences, based on two LLM — ChatGPT-4o and ChatGPT-4o-mini — and for system prompts at the three levels: A1, A1+, and A2.

## 2. Generative AI for Chinese as a foreign language learning within the CEFR framework defined by the EBCL Project

### 2.1 Related work

Much like tandem learning, which promotes peer interaction as developed by Hilton (2019), emphasizing real linguistic exchanges to foster mutual learning and language skill acquisition, chatbots can also facilitate interactive exchanges with immediate feedback, thereby creating a personalized tandem environment. Although using chatbots as an educational tool for language learning is not new (Eleni, 2020), the use of those based on large language models (LLMs) is relatively recent (Li et al., 2024). There is thus a blooming field of research on the use of LLMs for education (Cong, 2024; Xu, 2024), the specific use of ChatGPT (Zhao et al. 2024), and prompt engineering (Liu et al., 2023).

Many literature reviews, such as Huang (2020), are highly relevant but are from a pre-ChatGPT and pre-LLM era (for example, the term "LLM" itself is absent in that review). This literature review explores the use of chatbots in language learning, going beyond earlier studies that focused on specific aspects. The study analyzes their utility in first, second, and foreign language learning. By examining 25 empirical studies, three technological affordances (speed, ease of use, personalization) and five pedagogical uses (interlocutors, simulations, transmission, assistance, recommendations) were identified. Their conclusions show that chatbots foster social presence through affective and coherent communication. However, the authors note that challenges include technological limitations, novelty effects, and cognitive load. As mentioned above, since 2022, the landscape has changed, and we have excluded from our bibliography any related works that do not mention LLMs or predate these developments, except for those specifically related to Chinese (see Appendix 7). In this regard, in 2023, Mahlow (2023) concludes that "the integration of large language models and artificial intelligence in language learning and teaching in general—and in learning how to write in particular—has the potential to enhance the learning experience by providing new and innovative ways of learning" and that the technology now offers almost limitless possibilities, putting pedagogy and language teaching under pressure to adapt and leverage these new tools.

In terms of chatbots, ChatGPT (developed by OpenAI) is not the only one. Examples of other LLM-based chatbots include Claude (developed by Anthropic), Copilot (integrated into Microsoft's AI), HuggingChat (based on the Hugging Face platform), Gemini (formerly Bard, developed by Google), and Perplexity (developed by Perplexity AI). Other notable models include LLaMA (Meta), Ernie Bot (Baidu), or Mistral (Mistral) which enrich the LLM landscape with various applications, notably in the open-source and specialized domains. Nevertheless, ChatGPT has received the most publications, and its use for language learning has been systematically analyzed during its first year of existence, from November 2022 to November 2023 (Li et al., 2024). This analysis on ChatGPT and language education reveals a strong interdisciplinarity spanning computer science, psychology, and education. Most of the studies are empirical, focusing on higher education and ethical issues. This study shows that the main gaps in the 36 articles analyzed include the need for longitudinal studies and in-depth explorations of *learner perceptions* and *feedback quality*.

A recent in-depth analysis of 67 studies demonstrated that integrating AI chatbots into education offers numerous benefits for students and educators (Labadze et al., 2023). Students particularly benefit from homework assistance, a personalized learning experience, and the development of various skills. This article also highlights major challenges, notably concerns about reliability, accuracy—*hallucinations* (see Table 1)—and ethical issues surrounding AI use. The specific use of ChatGPT in learning, teaching, and assessment is also the subject of much research (Glaser et al., 2023). These studies also emphasize that, while making learning more personalized and effective, ethical concerns and limitations related to its impact on human interactions and the learning process remain. Glaser et al. (2023) specifically conclude that ChatGPT must be inclusive for all learners and address concerns about its effect on human interaction and critical thinking skills.

In the field of Chinese language learning, as mentioned in the introduction, several studies have focused on the use of ChatGPT in teaching Chinese (Zhao et al., 2024; Chen et al., 2023; Jiang et al., 2024; Li et al., 2024). For instance, in improving conversational learning and grammar practice (Chen et al., 2023), developing students' critical thinking through interactive scenarios and virtual debates (Jiang et al., 2024), and enabling differentiated instruction by tailoring exercises to the specific needs of students (Li et al., 2024). These studies rely on the use of prompts tailored to the tasks and focus on achieving learners' objectives. Our study focuses more on the prompts and ChatGPT's adherence to the constraints provided by the teacher. A literature review based on the search for academic articles with "chatbot" "learning" and "Chinese" in their titles identified other earlier studies, such as Chen et al. (2023), which evaluated the impact of a chatbot designed for Chinese vocabulary learning by comparing its effectiveness in two learning environments: in a classroom with multiple students and one-on-one with a tutor. This study involved 58 students divided into two groups. The results show that the chatbot significantly improved student performance, with better outcomes in one-on-one sessions compared to the classroom.

There are also several commercial applications that use chatbots for Chinese language learning, such as Duolingo (https://www.duolingo.com), HelloChinese (https://www.hellochinese.cc), LingoDeer (https://www.lingodeer.com), Super Chinese (https://www.superchinese.com), Tandem (https://www.tandem.net), Italki (https://www.italki.com), Busuu (https://www.busuu.com), Rosetta Stone (http://rosettastone.com), etc. Additionally, there are *assistants* (see Table 1) in ChatGPT, such as "Chinese Teacher", "Your Chinese Teacher", "Mandarin Chinese Teacher", "Chinese", etc., whose system prompts are not accessible. Other platforms, like HuggingChat (https://huggingface.co/chat/assistants), offer *assistants* where the system prompts are accessible. The first is called "CHINESE LANGUAGE LEARNING HELP», and its system prompt is: "You are a very experienced Chinese professor teaching Chinese; you will reply to any questions in Chinese characters, Chinese pinyin, and English translation, giving also communication and learning tips. Use any information you have access to". This assistant, mentioned in the introduction, uses the Mistral-Nemo-Instruct-2407 model which is a *fine-tuned* version (see Table 1) of the LLM Mistral-Nemo-Base-2407. Another assistant based on the same LLM is called "Commercial English and Chinese Expert". Its system prompt is: "You'll act as a professional expert in both commercial English and Chinese, assisting users in rewriting emails, letters, and contract terms".

All these assistants allow for written interactions in Chinese that are smooth, tailored to learners' needs, and enriched with pedagogical tips. Thanks to the accessibility of system prompts on platforms like HuggingChat, users can better understand how the assistant works and adjust their interactions accordingly. However, to benefit from the undeniable quality of the latest GPT-4 models, a user-provided prompt-based solution to ChatGPT is to create personalized learning scenarios. By providing a detailed *system prompt* structured around specific needs, users can design an assistant tailored to their level, with responses enriched by grammatical explanations, contextual usage examples, and tips to enhance active language practice (Li et al., 2024). This is the approach we have adopted in our study.

## 2.2 Specific challenges in teaching Chinese

Teaching Chinese presents unique challenges compared to alphabetic languages due to the logographic nature of its writing system—that is, teaching a second language with a non-phonographic writing system. The "character threshold" approach can enable a gradual and constructive program for learning Chinese writing.

### 2.2.1 An academic schism at the heart of Chinese Language Teaching

#### Dualistic pedagogical approach to Chinese

Unlike the teaching of other languages, where methods more or less share the same perspective on the language they aim to convey, Chinese didactics is fundamentally divided between two approaches. One treats Chinese as if it were, ultimately, a language like any other, viewing its non-alphabetic writing system as a matter of form rather than substance. The other immediately posits that teaching Chinese completely reshuffles the deck in terms of language didactics!

The first option, adopted by contemporary China in its methods for teaching Chinese as a foreign language since the 1950s, reflects a purely instrumental conception of Chinese writing. It treats the writing system as merely a different graphic system whose sole function is to record the language in writing, having no proper didactic impact. The most manifest consequence is the failure to provide the meanings of the characters composing a disyllabic word, thus denying the logographic nature of Chinese writing! For example, the word "Zhongguo" would be presented as an indivisible block meaning "China", without indicating that the two distinct characters composing it mean "middle" and "country", respectively, to denote the "Middle Kingdom". Within this option, characters are treated as a set of strokes or sometimes components, without considering their meanings or providing any memorization aids.

The second approach acknowledges the split reality of this language, with two minimal units—the word and the character—and two strata of the language: the linguistic stratum and the graphic stratum. In opposition to the first approach, which we term monistic, the dualistic approach in Chinese didactics recognizes the character as a semantic unit. This recognition dates back to the first institutional Chinese methods published in France in the 19th century, even considering it as a full-fledged minimal didactic unit by making frequency and lexical recurrence the criteria governing the choice and appearance of characters in the editorial design of textbooks, as is the case with the appearance of words (DeFrancis, 1966; Bellassen, 1989, 2018). Linguist Guo Shaoyu (1893–1984) distinguished as early as 1938 between "character-based written language as opposed to word-based spoken language" (Xu, 2008).

Swedish linguist Bernhard Karlgren emphasized in 1923 the particular relationship between spoken and written Chinese languages by identifying traditional literary language as a visual language rather than merely an auditory one (Karlgren, 1923/1930). Guo Shaoyu (1946/1985) notably suggests that this difference between spoken and written language, specific to Chinese, is not found in other languages. The article states: "In Chinese, there exists the special phenomenon of two languages, 'graphic language' and 'sound language,' which gives rise to many special laws not observed in languages other than Chinese" (Guo, 1946/1985).

### 2.2.2 Chinese facing the challenge of its "eurocompatibility" within the CEFR: the case of France

In the early 2000s, when foreign language curricula in French primary and secondary education underwent extensive renovation—including Chinese for the first time at the request of the Ministry of National Education—the CEFR had just been officially published in 2001. It was set to become the reference framework for 57 European countries, members of the Council of Europe (2001), encompassing some 800 million inhabitants. Chinese had to confront the CEFR's new approach to language teaching (Bellassen & Zhang, 2008; Bellassen, 2012; Zhang-Colin & Gianninoto, 2022; Lin-Zucker, 2024), particularly the redefinition of language activities meant to structure this teaching according to the proficiency levels (A1, A2, B1, B2, C1, C2). Since China did not wish to design the competency framework for Chinese under the guidance of the relevant European bodies, two unprecedented initiatives were taken in Europe: the first being the complete Chinese curricula published by the French Ministry of National Education in the early 2000s, and the second being the *European Benchmarking Chinese Language* (EBCL) project, undertaken between 2011 and 2013, explicitly dedicated to drafting a competency framework for Chinese aligned with the CEFR.

The officially published Chinese school curricula starting from 2002 inherited a decisive advancement with the conception and publication in 1985 of the SMIC (French acronym of "Seuil Minimum Indispensable de Caractères"), the first threshold aimed at programming 400 high-frequency characters. Indeed, this first sinographic threshold, based on criteria of frequency and combinatorial capacity of characters, not only recognizes the character as a full-fledged didactic unit but also provides an effective solution to the non-alphabetic nature of a language included in the French baccalaureate exams. Since the written exam subjects are proposed nationally, the texts offered in written exams finally had a clear and precise rule: to provide comprehension aids only when a word is composed of characters outside the threshold. With the historic publication of the Chinese school curricula from 2002 onward, the approach to written and sinographic competence in Chinese through character thresholds was fully completed with the official publication of different thresholds according to the curricular path in Chinese: a threshold of 805 characters for Language 1 (from 6th grade to 12th grade), a threshold of 505 characters for Language 2 (from 8th grade to 12th grade), and a threshold of 405 characters for Language 3 (from 10th grade to 12th grade). A final threshold of 1,555 characters intended for the intensive Chinese school path called "Section Internationale de Chinois" was published in 2008. Thus, the Chinese character was elevated to the status of a criterion for controlling written comprehension and expression, echoing the editorial principles of John DeFrancis's (1966) and Joël Bellassen's (1989) Chinese methods. The character thresholds constitute a prerequisite for performing written comprehension and expression activities. Published under the authority of the French Ministry of National Education, these thresholds officially express that Chinese stands out due to its linguistic and didactic uniqueness, as the sinographic thresholds preexist the definition of proficiency levels for written activities.

The second initiative is the EBCL project, which brought together experts from four European countries (England, France, Germany, Italy) and designed the first CEFR-adapted descriptors for Chinese. The work contributed to finalizing levels A1 and A2 and resumed this year after a 10-year hiatus to cover the other levels up to C2. The qualitatively central point of this foundational project is the principled demarcation from languages that include an effective link between grapheme and phoneme, whereas the logographic nature of Chinese writing introduces a break in this link. Therefore, the EBCL project posits a "graphemic competence" as a prerequisite, embodied through the consideration of Hanyu Pinyin phonetic transcription for Chinese users, as well as through the idea of character thresholds—in this case, a threshold of 320 characters at A1 level and another of 630 at A2 level. While the CEFR devoted only a few lines to orthographic competence, treating it relatively marginally, the EBCL project makes it a qualitatively fundamental point. It thus aligns with the dualistic approach to Chinese and carries within it the necessary dissociation of the progression of oral and written competences in Chinese, as was stated in the preamble to the 2002 French School Curricula. This affirms the unique nature of learning Chinese and, ultimately, a certain incommensurability between oral and written competences in Chinese.

### 2.3 AI facing the challenge of a split Chinese language

The potentials of artificial intelligence and certain of its avatars, including conversational robots or chatbots, have generated a leap forward in managing the split unique to Chinese between two minimal pedagogical units—the word and the character—and two strata of Chinese as a language of learning: linguistic and graphic, both intrinsically related and dissociated. We posit that an effective solution can be provided to this didactic conflict specific to Chinese, not only in terms of teaching and textbook design but now directly dedicated to learning itself. We know that one of the key words for effective learning is lexical

recurrence—the re-exposure and reuse of important words and newly learned words. Our proposal is to simultaneously cross lexical recurrence and sinographic recurrence with the help of a chatbot, that is, to couple the strategic principle of didactic reformulation in oral language (saying things with already learned words) and the recurrence of high-frequency characters belonging to a given sinographic threshold. In other words, ensuring respect for the frequency of the words we say but also for the characters used to write them. This involves applying the principle of economy (less is more) not only at the level of words within dialogued language interaction activities but also at the level of characters from given sinographic thresholds, thus providing an efficient response to this didactic conflict resulting from the duality of Chinese.

The envisaged language activity would be novel, engaging an interaction between a learner and a chatbot, further increasing both lexical and graphic recurrence. For example, it would take the following form:

筷子是什么？ ("What are chopsticks?")

1. Generated from a threshold of 200 frequent characters:

   筷子是中国人、日本人吃饭用的东西。

   ("Chopsticks are things that Chinese and Japanese people use to eat».)

2. Generated from a threshold of 400 frequent characters:

   中国人吃饭时用来拿住饭菜的长东西。西方人不怎么会用。

   ("A long object used by Chinese people to hold food when eating. Westerners don't really know how to use them».)

3. Generated from a threshold of 900 frequent characters:

   用木、竹等制成的细长的物体，中国用它把饭、菜等食物拿起来。

   ("A thin and long object made of wood, bamboo, etc.; in China, they use it to pick up rice, dishes, and other food».)

Such a mode of learner-chatbot interaction is a pedagogical modality oriented toward reformulation for lexical explanation. Another pedagogical modality would be the simulation of a natural dialogue, but also "conditioned" by a given sinographic threshold. In such a chatbot usage, the option of displaying pinyin vs. characters allows reflecting the induced differentiation from the oral/written-graphic split in Chinese. A new path addressing head-on the squaring of the circle in learning Chinese would then open...

### 3. To each student their own chatbot

The fact that a student can provide a different system prompt based on their level allows the learner to request that the exchanges use only the Chinese characters that fall within the character threshold they are aiming to acquire at their current stage of learning. It can also generate feedback corresponding to the learner's proficiency level and specific learning objectives. For example, preparation for an exam, improvement in conversational fluency, mastery of specialized vocabulary, and even the type of feedback the learner wishes to receive. Guided by the system prompt, the chatbot can then dynamically adapt its responses, exercises, and explanations to best meet the individual needs of the user. Flexibility in the chatbot's programming is crucial: it allows for fine-tuning parameters such as the difficulty level of interactions, the frequency and type of corrections, or the integration of cultural and contextual elements into the lessons. These adjustments can be made through an iterative process of model tuning, where learner feedback is used to continuously refine the chatbot's behavior. This personalized approach not only optimizes learning efficiency but also increases learner motivation and engagement by offering a tailored experience.

To implement this vision, several key concepts come into play, including: prompt engineering (the design of precise instructions to guide the AI's behavior), system prompts (basic instructions defining the chatbot's role and capabilities), generation parameters (which control aspects such as creativity or the precision of responses), and contextual adaptation (the model's ability to adjust in real-time to the learner's changing needs). By mastering these concepts and applying them judiciously, it becomes possible to create truly adaptive, user-centered language learning experiences. However, students are often unfamiliar with the use of prompts, as shown in the experiment conducted in class and described in subsection 3.2. At the end of this same section, we propose two prompts per level—A1, A1+, and A2—one relying on providing ChatGPT with an actual list of target characters for the level and the other without.

## 3.1 Feedback on prompt usage by "naïve" students

In a limited empirical study, we sought to understand how learners, without prior training in the use of precise prompts, interact spontaneously with ChatGPT in the context of their Chinese learning. These "naïve" users are in the exploratory phase of the tool and have not yet acquired the necessary skills to formulate optimized requests. This observation phase allows us to identify recurring patterns in prompt usage and better guide future users toward a more strategic approach.

The participants in this study are first-year university students with an A1 level in Chinese, according to the CEFR. Their goal is to reach A2 level by the end of the academic year. The group is mainly composed of 18- to 19-year-old students, all motivated by the desire to acquire functional proficiency in Chinese through written comprehension and expression exercises and the use of digital tools such as chatbots. Below, we present some questions that these "naïve" students of prompting technique posed to their chatbot following their Chinese teacher's instruction to study Chinese with ChatGPT. The teacher's instruction was as follows: "*You must work on general reading comprehension in Chinese (read and understand various texts as a whole, such as articles or short stories) using ChatGPT for this purpose. Send your teacher a copy of the questions you asked ChatGPT.*"

Here are the answers (i.e. user prompt) to this instruction that were sent by six of our students:

*Prompt used by Student 1:*
- "马丁下星期一有中国文学课。除了中国文学，马丁还上语法课、会话课和汉字课。上中国文学课的老师教书教得很好，学生们都觉得很有意思。马丁这个学期的功课很多，每次上课都很紧张。"
  Explain to me the grammar rules to know in this text.

Prompt used by Student 2:
- Convert the text to pinyin.
  What does 紧张 mean?

*Prompt used by Student 3:*
- I want you to help me understand this text:
  First, make me a list of new words to learn. Second, explain to me the important grammar points in this text. Then, make a list of questions about the text in Chinese as a practice exercise.

*Prompt used by Student 4:*
- What is the main idea of this text?
  What is the meaning of 紧张 and 然后?
  What are the main grammar points in this text?
  What is the most important vocabulary in this text?

*Prompt used by Student 5:*
- Can you provide the pinyin for each sentence?
- What does this character mean? (I ask this question in English for each character I don't understand because I find the translation more accurate than in French.)
- Then I translate the text on my own and ask ChatGPT: "Summarize the entire text in Chinese," and I compare my understanding of the text with the summary in Chinese before asking ChatGPT: "Can you translate the text into English?" This helps me determine if I understood the text well.

*Prompt used by Student 6:*
- Make me a list of vocabulary from the text with pinyin and the French translation.

Several key observations can be made about these students' interactions with ChatGPT:

- Vague request formulation: The learners, not yet fully aware of the tool's capabilities, often asked general or imprecise questions, leading to broad and sometimes poorly suited responses for their learning level.

- Lack of structured follow-up: The students did not necessarily follow a logical progression in their questions. This resulted in fragmented learning sessions where various topics were addressed without pedagogical continuity.

- Limited use of feedback capabilities: Although ChatGPT offers the possibility of asking for clarification or corrections, few students thought to use it to obtain personalized feedback or more in-depth explanations of their errors, often settling for the initial responses provided.

We observed that unguided prompts do not yield optimal results. This can be attributed to an ill-defined pedagogical context—whether it be the learner's level, learning objectives (target level, language skills to be developed, whether lexical, grammatical, related to reading comprehension or written expression), or the specific content of the lesson being studied by the learner. Consequently, ChatGPT's responses are based solely on probability calculations in the realm of natural language, without considering the learner's actual level at this stage of their learning. As a result, the tool frequently exceeds the learner's level, as if they were taking a C2-level Chinese course while they had only validated A1 and were aiming for A2, without the teacher adapting the language to the correct level. These observations lead us to consider several avenues for improvement to guide learners toward more effective prompt usage. By developing examples of specific prompts and educating users on how to refine them, we hope to facilitate better interaction with chatbots, particularly in learning a language as complex as Chinese. This feedback will serve as a basis for developing strategies to optimize prompts and improve Chinese language acquisition as a foreign language.

## 3.2 Language competencies in Chinese from the EBCL project

We observed that learners often ask simple and direct questions in their prompts, without particular optimization for obtaining better answers. These queries, which can be described as "naïve", are the most common. Students immediately ask for detailed grammatical explanations of their lesson to be learned. They also spontaneously ask questions about unknown vocabulary. As several studies on the use of ChatGPT in Chinese language classrooms have shown, there is value in providing teacher-prepared prompts (Jiang et al., 2024).

To test the different prompts within this study based on the EBCL project, we selected activities listed in Table 2, as they are related to written skills, whether for reading comprehension or written expression. Language learning activities primarily concern online written interaction skills, so we need to choose the activities defined by the EBCL project that specify language acts related to written production and reception: reading and writing. Below are the items from the EBCL project related to CEFR levels. They correspond to specific competencies in reading and written production.

Table 2

*Three of the ten EBCL items in relation to CEFR levels (full table given in Appendix 4)*

| EBCL Descriptors | Level A1 | Level A2 |
| --- | --- | --- |
| RW1: Overall Reading Comprehension | Can understand very short, simple texts a single phrase at a time, picking up familiar words and basic phrases and rereading as required. Can understand simple forms well enough to give basic personal details (e.g.,name, address, date of birth). | Can understand short, simple texts containing the highest frequency vocabulary. Can understand short simple messages and texts containing basic high frequency everyday vocabulary relating to areas of personal relevance or interest. (A2+) Can understand short narratives about everyday things dealing with topics which are familiar to me if the text is written in simple language. (A2+) |

| EBCL Descriptors | Level A1 | Level A2 |
|---|---|---|
| RW2: Reading Correspondence | Can understand simple written messages concerning appointments (time and date), e.g.in sms phone messages, emails etc.<br>Can understand short simple greetings and messages e.g. via SMS, on birthday cards and invitations. (A1+)<br>Can understand simple personal information (family, age, hobbies etc.) exchanged on social networking platforms. (A1+) | Can understand short simple personal letters giving or requesting information about everyday life or offering an invitation.<br>Can identify different types of letters relating familiar topics (information request, orders, complaints etc.) (A2+) |
| IW3: Notes, Messages & Forms | Can write numbers and dates, own name, nationality, place of birth/ residence, age, date of birth or arrival in the country etc. such as on a hotel registration form.<br>Can ask for or pass on personal details in written form.<br>Can write a short simple note or message (e.g., to tell somebody where I am or where to meet, to say that someone telephoned, arranging to meet someone, explaining absence). (A1+) | Can compose short, simple notes and messages relating to matters in areas of immediate need.<br>Can fill in a questionnaire giving an account of my educational background, my job, my interests and my specific skills.<br>Can take a short, simple message provided he/she can ask for repetition and reformulation about familial topics. (A2+) |

Each EBCL item aims to develop specific competencies, and the associated CEFR levels ensure a structured progression in acquiring written language, allowing learners to gradually master more complex tasks as they advance in their Chinese learning. Level A1 focuses on simple and straightforward tasks, with limited vocabulary and basic grammatical structures, while level A2 allows learners to carry out slightly more complex tasks, with more detail and a greater ability to express themselves.

## 3.3 Using system prompts as guidelines to align LLM responses with pedagogical objectives

The system prompts we propose to use (Tables 3 and 4) are not simple user prompts that ask a question. These are *system prompts* meaning prompts that condition the role played by the chatbot, in this case, ChatGPT, and impact all subsequent interactions (cf. Glossary Table 1). Constructing these system prompts requires what is known as prompt engineering (PengFei et al., 2023). There is an entire typology of prompts (cf. Appendix 1). We tried Few-Shot Prompts, One-Shot Prompts and Zero-Shot Prompt but ended up using a (cf. Appendix 1) Chain-of-Thought Prompt (CoT) which is represented by the four last line of the prompt of Table 3. below. This system prompt is the result of numerous tests, considering that the challenge of didactic prompt engineering lies in the fact that LLM (in this case, ChatGPT) do not strictly follow the given constraints (Zhang et al., 2024). One can experience this firsthand when asking ChatGPT to "provide a 100-word summary" of a text—it is unlikely that it will precisely adhere to this limit. This occurs because the LLM is instruction-tuned for tasks but remains an autoregressive language model (PengFei et al., 2023).

The proposed system prompts are intended to facilitate interaction around CEFR activities. What interested us was quantifying the extent to which the constraints were respected (instruction deviation) and whether the explicit addition of character lists from levels A1, A1+, and A2 had any impact. Additionally, we wanted to see whether the specific version of ChatGPT, whether ChatGPT-4o or ChatGPT-4o-mini, also had an effect.

Table 3

System prompt for A1 Student with *List* of characters

---

Role Instructions:  You will act as my Chinese tutor. First, introduce yourself and offer me a choice between three learning objectives.

You will ask the initial question in English, not Chinese:
- Revise a grammar point
- Have a conversation
- Understand and review a text

Alternatively, you may also offer an activity from the EBCL framework:
- RW1: Overall Reading Comprehension
- RW2: Reading Correspondence
- RW3: Reading for Orientation
- RW4: Reading for Information & Argument
- RW5: Reading Instructions
- PW1: Overall Written Production
- PW2: Creative Writing
- IW1: Overall Written Interaction
- IW2: Correspondence
- IW3: Notes, Messages & Forms

A1-level character list is: 爱八爸吧白百班半杯北本比笔边别病不菜茶长常车城吃出从打大到道的得地弟点电店东懂动都对多儿二饭方房飞非分父干刚高哥个给跟工公关馆贵国果过还孩海汉好号喝和很红后候花画话欢回会活火机几家间见叫姐今近进京九酒就觉开看可课口块快筷来老了累冷离里两六妈吗买卖忙么 没每美妹们米面名明母哪那男南难呢能你年您女朋票七期气汽前钱亲请去让人认日肉三山商上少谁什生师十时识 事难书水说思四岁他她太天听同外玩晚网为文问我五午西喜下先现想小些写谢心新信星姓兴学样要也一以意因影用友有雨语元远月运在再早怎这只知中重住字走昨坐作.

First answer to my request. When using Chinese words only use words that are made up exclusively of the A1-level characters provided in the list.  Second check if the characters you used are not present in the list of A1-level character. Third rephrase your responses to stay within the constraint of using A1-level characters.

---

Table 4

System prompt for A1 Student *Without* List of Characters

---

Role Instructions: You will act as my Chinese tutor. First, introduce yourself and offer me a choice between three learning objectives.

You will ask the initial question in English, not Chinese:
- Revise a grammar point

---

- Have a conversation
- Understand and review a text

Alternatively, you may also offer an activity from the EBCL framework:

- RW1: Overall Reading Comprehension
- RW2: Reading Correspondence
- RW3: Reading for Orientation
- RW4: Reading for Information & Argument
- RW5: Reading Instructions
- PW1: Overall Written Production
- PW2: Creative Writing
- IW1: Overall Written Interaction
- IW2: Correspondence
- IW3: Notes, Messages & Forms

First answer to my request. When using Chinese words only use words that are made up exclusively of the A1-level characters provided in the list.
Second check if the characters you used are not present in the list of A1-level characters.
Third, rephrase your responses to stay within the constraint of using A1-level characters.

The system prompts for levels A1+ and A2 with the list of characters are given in Appendix 2 and Appendix 3.

## 4. Comparative analysis of responses generated by ChatGPT

Any study on the use of ChatGPT in Chinese language classes yields results that, as noted by Li et al. (2024), are highly dependent on the prompts and provide different outcomes. They emphasize that as an AI model, ChatGPT is particularly responsive to instructions, and variations in these instructions can lead to vastly different responses. Therefore, a thorough understanding of the effectiveness of using ChatGPT for pedagogical purposes requires examining the impact of different prompts (Zhao, 2024).

### 4.1 Method

To evaluate the performance of two language models, "GPT-4o-mini" and "GPT-4o", we conducted a series of experiments using python programs accessing the ChatGPT Application Program Interface (API) (Ekin, 2023). These two models are optimized versions of the GPT-4 model, with "GPT-4o-mini" being a lighter and faster variant, while "GPT-4o" is designed for more complex tasks with greater processing capacity. In practice, this means that we developed a program capable of automatically querying ChatGPT thousands of times, simulating the responses that a student at the A1, A1+ or A2 level would provide to a system prompt. This allows us to test the models intensively and measure their performance under real-world usage conditions. The models were tested in conditions simulating those of a student, with a *temperature* set to 0.7 (see Appendix1).

We examined the impact of the presence or absence of an explicit list of characters in the system prompt across three linguistic proficiency levels ("A1", "A1plus", "A2") and ten distinct tasks: "RW1", "RW2", "RW3", "RW4", "RW5", "PW1", "PW2", "IW1", "IW2", and "IW3". Each combination of model, level, task, and prompt condition was repeated ten times to ensure the statistical robustness of the results obtained. This represents six hundred requests to ChatGPT-4o-mini and ChatGPT-4o, which cost a few cents and a few dollars, respectively, in total.

For each API call, the responses generated by ChatGPT were analyzed to identify the percentage of characters that did not belong to the specific character list defined for each linguistic level. This analysis allowed us to assess the extent to which the models adhered to the linguistic constraints associated with levels "A1", "A1plus", and "A2". By quantifying the occurrences of out-of-list characters, we were able to measure the models' fidelity in producing text that conforms to the requirements of each proficiency level. The percentage of out-of-list characters for each level, along with its standard deviation, is provided. The results are presented in Fig. 1a and Fig. 1b below for the use of ChatGPT-4o and ChatGPT-4o-mini, respectively (exact values are provided in Appendix 5 and 6 respectively).

Figure 1

*Results of the gain brought by the presence of the list in the system prompt, reducing the number of out-of-list characters for the corresponding Level.*
*a) left when using the GPT-4o Model*             *b) right when using the GPT-4o-mini.*

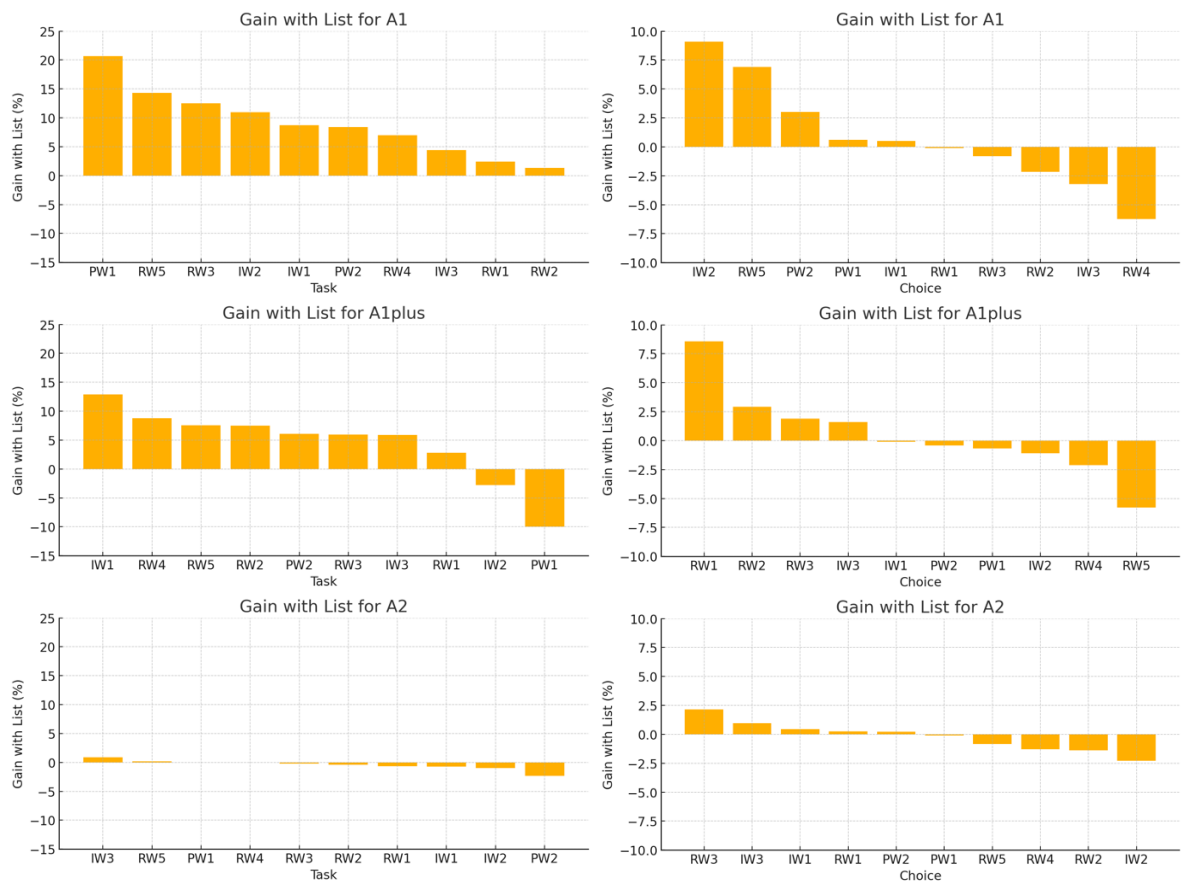

## 4.2 Discussion of statistical results on the structure of ChatGPT's responses

Let us first consider the ChatGPT-4o model, which is more advanced than ChatGPT-4o-mini. The addition of a character list specific to each level significantly improves performance at A1 and A1+ levels by reducing error percentages (first two bar charts of Fig. 1a). However, at the A2 level, the addition of the list leads to a negligible increase in the error rate, although the overall error percentages remain low in both scenarios. In contrast, with ChatGPT-4o-mini, the analysis shows that the addition of a character list does not significantly reduce the percentage of generation errors, as the average differences in errors between the two conditions are minimal and within the margin of the provided standard deviations. Thus, adding a character list does not improve generation performance in terms of error reduction with ChatGPT-4o-mini, and it may not be necessary to include it for this purpose. However, incorporating a list in tasks for A1 and A1+ levels improves accuracy, whereas at the A2 level, it does not present significant benefits and may be considered optional.

## 4.3 Discussion of results from the perspective of prompt content

Beyond the relevance of whether or not to include the Chinese character threshold in the prompt, from a pedagogical perspective, ChatGPT's output on the content provided to the learner demonstrates a certain quality. Below in Table 5, we cite an example of the result from a prompt with the "RW2" item from the EBCL project, which relates to the language act aimed at developing the 'Reading Correspondence' competence at the A1 level. This is not a caption of a real interaction as the result was obtained automatically using the ChatGPT API (cf. Methods described in §4.1).

Table 5

*The answer was obtained using the ChatGPT API with the model GPT4o, the prompt from Table 3 and the response 'RW2'. The output sentence of the python program 'Level A1 Choice RW2 Run 3/10' indicates that this task is at the A1 language proficiency level, focusing on a choice "RW2" EBCL activity. It also indicates that this is the third iteration out of a total of ten attempts (cf. section §4.1).*

| Level A1 Choice RW2 Run 3/10 |
| --- |

Hello! My name is [Your Tutor's Name], and I'm here to help you learn Chinese. Today, we will focus on "Reading Correspondence" (RW2), which involves understanding and interpreting written communication, such as emails or letters. Let's start with a simple email in Chinese. I'll provide you with a basic email, and then we can go over it together.

---
Subject: 你好
亲爱的朋友，
你好！我想告诉你，我今天很高兴，因为我学习了很多新的汉字。我希望你也有一个愉快的一天。
祝好，
小明
---
Please read the email and then answer the following questions:
1. Who is the email addressed to?
2. Why is the sender happy today?
3. What does the sender wish for the recipient?
Take your time to read and understand the email. Once you're ready, we can discuss your answers.

Let us first recall that an A1-level learner is expected, regarding reading comprehension, to "understand simple, short messages in emails". Among the characters used in the short email created by chatGPT above, only one character —"愉"— does not belong to the A1-level threshold of 320 characters. It is the only character used that does not meet the EBCL project's requirements for A1 level. One might wonder why ChatGPT did not use the word 快乐 instead of 愉快. The former is roughly synonymous with the latter and contains the character 乐, which is part of the 320 characters required at the A1 level.

Despite this discrepancy, the format of the email, the clarity of the instructions, and the pedagogical relevance of the questions posed about the text justify the effectiveness of ChatGPT's proposed approach. The learner is guided step by step through the exercise, with questions directly related to the message's content, thereby fostering progressive understanding and active learning. Furthermore, this type of interaction allows the learner to deepen their reading comprehension skills while staying close to real communication situations, such as reading emails. This immersive and interactive format provides significant pedagogical value, even though minor adjustments could be made to fully align the content with the specific requirements of the EBCL project.

## 4.4 Limitations and ethical considerations

Oral interaction with chatbots is limited by the very nature of the tool, which relies primarily on a text-based interface. Unlike human interaction or advanced voice recognition systems, chatbots based on large language models (LLMs) such as ChatGPT do not allow for true oral conversation. This can make the aspect of oral production less intuitive than written recognition or production. A study was conducted to compare the effectiveness of a virtual agent and a physical robot named Qilin in helping English-speaking adults learn spoken Mandarin (Hargis, 2022). Participants interacted with a bilingual chatbot using three modes:

translation, chat, and quiz. The results showed that the physical embodiment of the robot fostered higher engagement and reduced discouragement, despite having the same vocal features as the virtual agent. Difficulties related to Mandarin tones and polyphonic characters contributed to the learners' discouragement. The study highlights the potential of the robot to improve Mandarin learning for English speakers.

Although ChatGPT offers the possibility of reading its responses aloud and speaking to it in Chinese, it does not have the features of commercial platforms dedicated to learning Chinese. Moreover, ChatGPT does not retain true memory when a new interaction is initiated. Additionally, each time a specific educational prompt is needed, it must be copied and pasted. One solution to this problem would be to create a personalized assistant that could be integrated into an educational platform or workspace. This assistant could be programmed to retain a history of interactions and the student's learning preferences, allowing for continuity in Chinese learning sessions. It could also adapt prompts and exercises based on the student's progress, offering continuous personalization without the need to reintroduce the same information in each session. Additionally, by using APIs and content management systems, this assistant could store educational resources such as vocabulary lists or sentence models, facilitating quick access to relevant materials.

However, these limitations do not mean that ChatGPT is useless for practicing oral production. Even though the chatbot does not speak or listen to the learner, it remains a powerful tool for structuring and guiding oral exercises. The learner can use the prompts provided by the chatbot to repeat sentences aloud, thereby reinforcing their pronunciation and fluency skills. Moreover, by providing phonetic transcriptions and detailed explanations of tones and nuances of the language, the chatbot creates an environment in which the learner is encouraged to actively verbalize the generated content.

This approach can be enhanced by external tools, such as speech synthesizers or speech recognition systems, which can extend the LLM's capabilities in terms of oral production. In fact, the ChatGPT applications on smartphones can provide voice input and output features, enabling users to interact with the chatbot through spoken language, further enhancing the learning experience. The learner can thus benefit from indirect feedback or increased exposure to the target language while maintaining control over their learning pace. By combining these different approaches, LLMs can still play an important role in improving oral skills, particularly in terms of repetition, interactive practice, and correction based on comparison with reliable linguistic models.

Using ChatGPT as a tutor for learning Chinese, particularly in writing tasks, raises several ethical considerations that are not specific to Chinese (Li, B., 2024; Jeon, 2023; Glaser, 2023; Imran, 2023; Labadze, 2023). While this usage allows students to access a virtual tutor available at all times, the excessive reliance on its use, the question of the authenticity of written productions, and the risk of limiting students' creativity and autonomy are of concerns. Additionally, there is the issue of data privacy, as students' personal information and written outputs could be exploited or stored by the model, which requires careful consideration of privacy protection in an educational context. This also raises the question of academic integrity, as students might not fully develop their independent writing skills. Moreover, ChatGPT's ability to provide precise grammatical or lexical suggestions can lead to an oversimplification of learning, without accounting for the cultural and contextual nuances that are essential to mastering a language. Finally, the fact that the most powerful models (in this case, ChatGPT-4o) are superior may lead to inequality among students, with wealthier individuals having access to a more performant system. Teachers must, therefore, supervise the use of these tools, ensure that students actively engage in their learning, and use ChatGPT as a complement rather than a substitute for autonomous learning (Crawford et al., 2023; Vaccino-Salvadore, 2023; Adel et al., 2024; Cao et al., 2024).

## 5. Conclusion

Chatbots based on LLMs, accessible via the internet or dedicated applications, make language learning more interactive and engaging today. By simulating natural conversations and creating interactive scenarios with instant feedback, they surpass traditional passive learning methods. Our study highlighted how, through well-structured *system prompts*, learners of Chinese as a second language can benefit from personalized interactions that adjust to their level, following the recommendations of the *Common European Framework of Reference for Languages* (CEFR) and using the character lists defined by the *European Benchmarking Chinese Language* (EBCL) project. These prompts allow for controlling and calibrating linguistic productions based on character frequency, promoting progressive and controlled exposure to the most essential sinographic characters.

However, we have experimentally shown that chatbots like ChatGPT do not systematically follow the instructions provided by prompts, even though more advanced versions like ChatGPT-4o outperform lighter models like ChatGPT-4o-mini. We showed that adding a character list improves accuracy in EBCL activities at the A1 and A1+ levels, helping learners at those stages. However, for A2-level tasks, the list does not provide significant benefits and can be considered optional, as it does not enhance error reduction or overall instruction deviation when using models like ChatGPT-4o-mini.

These results underscores the importance for teachers to develop appropriate prompts and continue improving them to ensure their effectiveness. Future research includes automatically improving prompts (Pryzant et al., 2023) and creating *assistants* based on these optimized prompt to make them widely accessible. Tests could also be conducted to evaluate the efficiency of the provided prompts when used with other LLM-based chatbots such as LLAMA, Mistral, or Qwen, ensuring broader applicability across various large language models. The use of generative AI for Chinese language learning also has the advantage of bridging the gap between oral and written skills, a crucial challenge in teaching this logographic language. By adjusting prompts, chatbots can provide exercises on character recognition and oral production, particularly via Pinyin, thus creating a synergy between these two skills.

Personalized chatbots powered by generative AI have the potential to revolutionize language learning, especially for non-alphabetic languages like Chinese. By providing adapted system prompts tailored to individual learners' needs and levels, these chatbots can maximize appropriate exposure to the target language. However, continuous evaluation is essential to assess their effectiveness and to optimize their integration into educational contexts. Educators should remain actively involved in developing and refining prompts to fully harness the benefits of this technology.

## References

Adamopoulou, E. & Moussiades, L. (2020). Chatbots: History, technology, and applications. Mach. Learn. Appl. 2, 100006.

Adel, A., Ahsan, A. & Davison, C. (2024) ChatGPT Promises and Challenges in Education: Computational and Ethical Perspectives. *Educ. Sci.* 14, 814.

Bellassen, J. (1989). *Méthode d'initiation à la langue et à l'écriture chinoises*. La Compagnie.

Bellassen, J. (2009). *Snowballing Chinese*. Sinolingua Press.

Bellassen, J. (2012). « Is Chinese eurocompatible? Is the Common Framework Common? The Common European Framework of Reference for Languages facing Distant Languages (Keynote Report) », International Symposium Report 2011, Tokyo University of Foreign Studies, 23-33.

Bellassen, J. (2018). 一元论抑或二元论:汉语二语教学本体认识论的根本分歧与障碍 [Monism vs Dualism: The Fundamental Divergency and Obstacle in the Ontological Epistemology of Chinese Second Language Teaching]. 华文教学与研究 [TCSOL Studies], No. 4, Sum No. 72.

Bellassen, J. & Zhang, L. (2008). 《欧洲语言共同参考框架》新理念对汉语教学的启示与推动 [The enlightenment and the impetus of the new approach of the Common European Framework of Reference for Language on the Chinese language Teaching]. 世界汉语教学 [Chinese Teaching in the World], 3.

Bellassen, J. (2024). 再说中文第二语言文字中的分裂问题：国际中文教育史上"缺失之环"—德范克 [The Schism in Teaching of Chinese as a Second Language: The "Missing Ring in the History of Chinese Language Education—John DeFrancis], 全球中文发展研究 [Journal of Global Chinese Language Development], 2.

Cao, Y., Fan, J. & Yang, Q. (2024) ChatGPT in Education: Ethical Predicaments of Generative AI. *Trans. Soc. Sci., Educ. Humanit. Res.* 11, 64–69.

Chen, J. L. R. J.-H., Ren, X., Jiang, X. & Chen, C.-H. (2023). Exploring the Use of ChatGPT in Chinese Language Classrooms. Int. J. Chin. Lang. Teach. doi:10.46451/ijclt.20230303.

Cong, Y. (2024). AI Language Models: An Opportunity to Enhance Language Learning. Informatics 11, 49.

Council of Europe (2001), *The Common European Framework of Reference for Languages: Learning, Teaching, Assessment.* Cambridge University Press.

Crawford, J., Cowling, M. & Allen, K.-A. (2023) Leadership is needed for ethical ChatGPT: Character, assessment, and learning using artificial intelligence (AI). *J. Univ. Teach. Learn. Pr.* 20.

DeFrancis, J., Teng, C. Y. Y., & Yung, C. S. (1966). *Beginning chinese reader*. Yale University Press.

Ekin, S. (2023) Prompt Engineering For ChatGPT: A Quick Guide To Techniques, Tips, And Best Practices. *Authorea Prepr.* doi:10.36227/techrxiv.22683919.v2.

Glaser, N. (2023). Exploring the Potential of ChatGPT as an Educational Technology: An Emerging Technology Report. Technol., Knowl. Learn. 28, 1945–1952.

Guder, A. (2014). Reading Competence and Graphemic Competence: Impacts of the Chinese Writing System on Designing Competence Descriptors for the "European Benchmarks for the Chinese Language" (EBCL) Project, CHUN NR. 29/2014, S. 5–29.

Guder, A., & Chinesisch eV, F. (2015). European Benchmarks for the Chinese Language (EBCL): Levels A1. 1/A1/A1+/A2/A2+. https://www.ouhanhui.eu/ebcl/

Guo, S Y. (1946/1985). 中国语言所受到的文字的牵制 [The Constraint of Writing on the Chinese Language], 照隅室语言文字论集 [Collected Essays on Language and Writing from the Zhaoyu Room], 上海古籍出版社 [Shanghai Classics Publishing House].

Hilton, H. E. (2019). A psycholinguistic perspective on "tandem learning" in the foreign language classroom. In *Redefining tandem language and culture learning in higher education* (pp. 47-59). Routledge.

Huang, W., Hew, K. F. & Fryer, L. K. (2022). Chatbots for language learning—Are they really useful? A systematic review of chatbot-supported language learning. J. Comput. Assist. Learn. 38, 237–257.

Hugging Face. (2023). HuggingChat: An open-source AI ChatBot. Hugging Face. https://huggingface.co/chat

Hutson, M. (2024) Forget ChatGPT: why researchers now run small AIs on their laptops. *Nature* 633, 728–729.

Imran, M. & Almusharraf, N. (2023). Analyzing the role of ChatGPT as a writing assistant at higher education level: A systematic review of the literature. Contemp. Educ. Technol. 15, ep464.

Jeon, J. & Lee, S. (2023). Large language models in education: A focus on the complementary relationship between human teachers and ChatGPT. Educ. Inf. Technol. 28, 15873–15892.

Jiang, X., Li, J. & Chen, C-H. (2024). Enhancing Critical Thinking Skills with ChatGPT-Powered Activities in Chinese Language Classrooms. Int. J. Chin. Lang. Teach. doi:10.46451/ijclt.20240105.

Kalyan, K. S. (2024). A survey of GPT-3 family large language models including ChatGPT and GPT-4. Nat. Lang. Process. J. 6, 100048.

Karlgren, B. (1923/1930). 汉语的读音和符号 [Pronunciation and Symbols of Chinese]. 世界语言文化手册系列 [World Language and Culture Handbook Series].

Labadze, L., Grigolia, M. & Machaidze, L. (2023). Role of AI chatbots in education: systematic literature review. Int. J. Educ. Technol. High. Educ. 20, 56.

Li, B., Lowell, V. L., Wang, C. & Li, X. (2024). A systematic review of the first year of publications on ChatGPT and language education: Examining research on ChatGPT's use in language learning and teaching. Comput. Educ.: Artif. Intell. 7, 100266.

Li, Q., Zhang, J., & Cai, W. (2024). Utilizing ChatGPT to Implement Differentiated Instruction. *International Journal of Chinese Language Teaching*, 5(1). https://doi.org/10.46451/ijclt.20240106

Li, L., Zhang, H P., Li, C J., You, H W. & Cui, W Y. (2023). Evaluation on ChatGPT for Chinese Language Understanding. *Data Intelligence*. 5 (4): 885–903. doi: https://doi.org/10.1162/dint_a_00232

Liu, P., Yuan, W., Fu, J., Jiang, Z., Hayashi, H., & Neubig, G. (2023). Pre-train, prompt, and predict: A systematic survey of prompting methods in natural language processing. *ACM Computing Surveys*, 55(9), 1-35.

Lin-Zucker, M. (2024). CEFR 在法國漢語教學中的挑戰 [Challenges of Using the CEFR in Chinese Language Teaching: the case of France]. 華語學刊 [*Journal of Teaching Chinese as a Second Language* (TCSL)].

Mahlow, C. (2023). Large Language Models and Artificial Intelligence, the End of (Language) Learning as we Know it—or not quite? doi:10.35542/osf.io/da2rm.

Microsoft. (2023). Introducing the new Bing: AI-powered search engine. Microsoft. https://www.microsoft.com/en-us/bing

OpenAI. (2023). ChatGPT: Optimizing language models for dialogue. OpenAI. https://openai.com/research/chatgpt

Perplexity AI. (2023). Perplexity: AI-powered answer engine. Perplexity AI. https://www.perplexity.ai

Pryzant, R., Iter, D., Li, J., Lee, Y., Zhu, C., & Zeng, M. (2023). Automatic Prompt Optimization with "Gradient Descent" and Beam Search. Proceedings of the *2023 Conference on Empirical Methods in Natural Language Processing*, 7957-7968. https://doi.org/10.18653/v1/2023.emnlp-main.494

Vaccino-Salvadore, S. (2023) Exploring the Ethical Dimensions of Using ChatGPT in Language Learning and Beyond. *Languages* 8, 191.

Wallace, R.S. (2009). The Anatomy of A.L.I.C.E.. In: Epstein, R., Roberts, G., Beber, G. (Eds.). *Parsing the turing test: philosophical and methodological issues in the quest for the thinking computer.* Springer. https://doi.org/10.1007/978-1-4020-6710-5_13

Wang, F., Cheung, A. C. K., Neitzel, A. J. & Chai, C. S. (2024). Does Chatting with Chatbots Improve Language Learning Performance? A Meta-Analysis of Chatbot-Assisted Language Learning. Rev. Educ. Res. doi:10.3102/00346543241255621.

Weizenbaum, J. (1966). ELIZA—a computer program for the study of natural language communication between man and machine. Communications of the ACM, 9(1), 36-45. doi: 10.1145/365153.365168. Commun. ACM 9, 36–45.

Xu, T Q. (2008). 汉语字本位语法导论 [Introduction to Character-based Grammar of Chinese], 山东教育出版社 [Shandong Education Press].

Xu, S., Qin, L., Chen, T., Zha, Z., Qiu, B., & Wang, W. (2024). Large Language Model based Situational Dialogues for Second Language Learning. arXiv preprint arXiv:2403.20005. doi:10.48550/arxiv.2403.20005.

Zhang-Colin, Y., & Gianninoto, M. (2022). Les idées didactiques en chinois langue étrangère (CLE) depuis 1945: recoupements et croisements avec l'histoire du français langue étrangère (FLE). [Pedagogical Approaches in Chinese as a Foreign Language (CFL) since 1945: Overlaps and Intersections with the History of French as a Foreign Language (FFL).] *Documents pour l'histoire du français langue étrangère ou seconde*, (68).

Zhang, T., Shen, Y., Luo, W., Zhang, Y., Liang, H., Zhang, T., Yang, F., Lin, M., Qiao, Y., Chen, W., Cui, B., Zhang, W., & Zhou, Z. (2024). CFBench: A Comprehensive Constraints-Following Benchmark for LLMs. arXiv. https://doi.org/10.48550/arxiv.2408.01122

Zhao, Q., Hsu, Y. Y., & Huang, C. R. (2024). Large language model and Chinese near synonyms: designing prompts for online CFL learners. *Journal of technology and Chinese language teaching*, 15(1), 49-69.

# Appendix

## Appendix 1

**Glossary of LLM and prompt related terms**

| Term | Definition |
|------|------------|
| *User Prompt* | Text or question entered by the user to request a response from the model. It is the most direct and explicit form of prompt. |
| *System Prompt* | See. Table 1. |
| *Few-Shot Prompt* | A prompt that provides several examples of input-output pairs before asking the model to respond to a similar new task. |
| *Zero-Shot Prompt* | No example is given before the task. The model must provide a response without explicit guidance or context. |
| *One-Shot Prompt* | Only one example is provided to help the model understand the task before posing the question it must answer. |
| *Instruction Prompt* | An explicit directive is given to the model to accomplish a specific task, such as summarizing a text or providing an explanation. |
| *Chain-of-Thought Prompt (CoT)* | The model is guided to detail its reasoning step-by-step before providing an answer, to improve accuracy. The end of the prompts of Appendix 2 and Appendix 3 are example of CoT prompts. |
| *LLM Response Temperature* | The temperature controls creativity and variability in responses. Low temperatures (T~0) lead to deterministic answers; high temperatures (>1) increase creativity. |

## Appendix 2

Prompt for A1+ Student with *List* of characters

Role Instructions: You will act as my Chinese tutor. First, introduce yourself and offer me a choice between three learning objectives.
You will ask the initial question in English, not Chinese:
- Revise a grammar point
- Have a conversation
- Understand and review a text

Alternatively, you may also offer an activity from the EBCL framework:
- RW1: Overall Reading Comprehension
- RW2: Reading Correspondence
- RW3: Reading for Orientation
- RW4: Reading for Information & Argument
- RW5: Reading Instructions
- PW1: Overall Written Production
- PW2: Creative Writing
- IW1: Overall Written Interaction

- IW2: Correspondence
- IW3: Notes, Messages & Forms

A1+ level character list is: 爱八爸吧白百班半杯北本比笔边别病不菜茶长常车城吃出从打大到道的得地弟点电店东懂动都对多儿二饭方房飞非 分父干刚高哥个给跟工公关馆贵国果过还孩海汉好号喝和 很红后候花画话欢回会活火机几家间见叫姐今近进京九酒 就觉开看可口块快筷来老了累冷离里两六妈吗买卖忙么 没每美妹们门们米面名明母哪那男南难呢能你年您女朋票七 期气汽前钱亲请去让人认日肉三山商上少谁什生师十时识 事是市书水说思四岁他她太天听同外玩晚网为文问我五午 西喜下先想现小些写谢心新信星姓兴学样要也一以意因影 用友有雨语元远月这在再早怎这只知中重住子字走昨坐作 安包部差场床次村错第饿发法歌共狗黑或级鸡介零路 马慢猫脑 牛农旁片骑起千球热绍视手睡所铁头物息习系行休羊医音英右鱼园 院乐 云找者址祝自足最左做

First answer to my request. When using Chinese words only use words that are made up exclusively of the A1+ level characters provided in the list. Second check if the characters you used are not present in the list of A1+ level character. Third rephrase your responses to stay within the constraint of using A1+ level characters.

## Appendix 3

Prompt for A2 Student with *List* of characters

Role Instructions: You will act as my Chinese tutor. First, introduce yourself and offer me a choice between three learning objectives.

You will ask the initial question in English, not Chinese:
- Revise a grammar point
- Have a conversation
- Understand and review a text

Alternatively, you may also offer an activity from the EBCL framework:
- RW1: Overall Reading Comprehension
- RW2: Reading Correspondence
- RW3: Reading for Orientation
- RW4: Reading for Information & Argument
- RW5: Reading Instructions
- PW1: Overall Written Production
- PW2: Creative Writing
- IW1: Overall Written Interaction
- IW2: Correspondence
- IW3: Notes, Messages & Forms

A2-level character list is: 啊爱安八吧把爸白百班般办半帮报抱杯北备被本比笔毕边便变表别病博不步部才菜参餐层茶差长常厂场唱超车成城吃出初除厨楚穿传床春词此次从村错打大代带待单但蛋当到道得的灯等低地弟第典点电店定订丢东冬懂动都读独度短对多饿儿而二发法反饭方房放飞非费份风封服附父复该改干感刚高告哥歌格个给跟更工公共狗古故拐怪关观馆惯广贵国果过孩海汉好号喝和河贺黑很红后候湖护花化画话坏欢还换黄回会婚活火或机鸡级极急几己记际济继寄加家假价间

见件健江讲酱交饺叫铰教接街节结姐解介界借今斤金禁近进京经景净静九久酒旧就
局句决觉卡开看康考可渴刻客课空口裤块快筷拉来篮老乐了累冷离礼李里理力历丽
联凉两亮谅辆聊林零六楼路录旅妈马码吗买卖满慢忙猫毛贸么没每美妹门们米面免
民名明末母目拿哪那奶男南难脑呢能你年念鸟您牛 农努女暖欧怕乒乓旁胖跑朋片漂票
乒平七期其奇骑起气汽千前签钱 欢且亲轻情请秋球取去趣全然让热人认日肉如
赛三散色山商上少绍社身生声胖师十什时识实史始世市事视试室是收手首书术树双
谁水睡说司思死四送诉算虽岁孙所他她它台太谈汤堂套特踢提题体天填条铁听厅庭
通同头图外玩完晚万王网往忘望卫为位文 问我卧无五午务勿物西吸希息习洗喜戏系
下夏先现相想向像消小 校笑些写谢心新信兴星行姓性兄休修需许续选学雪牙亚烟言
羊阳 样药要也业夜一艺衣医已以易意因音印银应英影硬泳用邮油游友 有又右鱼愉
雨语元员园原远院愿月越云运咱再在早怎站张找照者 这真正证知只之直址止纸至治
中钟种重周洲主住助祝注专准桌着 子字自总走租足最昨左作坐做座

First answer to my request. When using Chinese words only use words that are made up exclusively of the A2-level characters provided in the list. Second check if the characters you used are not present in the list of A2-level character. Third rephrase your responses to stay within the constraint of using A2-level characters.

## Appendix 4

*The ten EBCL project items in relation to CEFR levels*

| EBCL Descriptors | Level A1 | Level A2 |
| --- | --- | --- |
| RW1: Overall Reading Comprehension | Can understand very short, simple texts a single phrase at a time, picking up familiar words and basic phrases and rereading as required.<br>Can understand simple forms well enough to give basic personal details (e.g.,name, address, date of birth). | Can understand short, simple texts containing the highest frequency vocabulary.<br>Can understand short simple messages and texts containing basic high frequency everyday vocabulary relating to areas of personal relevance or interest. (A2+)<br>Can understand short narratives about everyday things dealing with topics which are familiar to me if the text is written in simple language. (A2+) |
| RW2: Reading Correspondence | Can understand simple written messages concerning appointments (time and date), e.g.in sms phone messages, emails etc.<br>Can understand short simple greetings and messages e.g. via SMS, on birthday cards and invitations. (A1+)<br>Can understand simple personal information (family, age, hobbies etc.) exchanged on social networking platforms. (A1+) | Can understand short simple personal letters giving or requesting information about everyday life or offering an invitation.<br>Can identify different types of letters relating familiar topics (information request, orders, complaints etc.) (A2+) |
| RW3: Reading for Orientation | Can find time and price related information encountered in public places (price tags, opening hours).<br>Can understand single words and some very short phrases frequently encountered in public places on | Can find specific predictable information in simple everyday material such as advertisements, timetables, menus, directories, brochures.<br>Can skim small advertisements in newspapers, locate the heading or column I |

| EBCL Descriptors | Level A1 | Level A2 |
|---|---|---|
| | signboards, posters and brochures (e.g. "no smoking", "private", "station", "car park", "no parking", "keep left"), or in the classroom (e.g. "book", "very good"). (A1+) Can understand the most important menu items/options in a computer programme, such as "PRINT", "SAVE", "COPY". (A1+) | want and identify the most important pieces of information (price and size of apartments, cars, and computers). Can find specific, predictable information in simple everyday material such as advertisements, prospectuses, menus, reference lists and timetables. (A2+) Can understand everyday signs and notices in public places, such as streets, restaurants, railway stations; in workplaces, such as directions, instructions, hazard warnings. (A2+) |
| RW4: Reading for Information & Argument | Can get an idea of the content of simpler informational material and short simple descriptions, especially if there is visual support. | Can identify specific information in simpler written material he/she encounters such as letters, brochures and short newspaper articles describing events if pictorial support is provided and rereading permitted. Can identify important information in news summaries or simple newspaper articles in which numbers play an important role and which are clearly structured and illustrated. (A2+) |
| RW5: Reading Instructions | Can follow simple instructions in textbooks (listen / read / speak with classmate / note down). Can follow short, simple written directions if familiar with relevant proper names in their sino-graphemic form. (A1+) | Can follow clear, step- by-step instructtions on equipment encountered in everyday life such as simple food preparation instructions (e.g. instant noodles), public telephone, taking out cash or buying a drink from a machine mainly relying on visual support. Can understand regulations, for example safety, when expressed in simple language. (A2+) |
| PW1: Overall Written Production | Can write simple isolated phrases and sentences. Can write simple isolated phrases and sentences about some personal information (name, age, birthday, nationality, where I am, my family). Can write in short simple sentences a blog entry or a similar piece of writing introducing oneself with basic information such as family, jobs, hobbies, one's hometown, etc. (A1+) | Can compose a series of simple phrases and sentences connected together or linked with simple connectors like "but" and "because". Can compose simple connected texts on a range of topics within my field of interest. (A2+) Can compose in short simple sentences a blog entry or a similar piece of writing about one's im- pressions of something one just bought (such as a digital camera) or of an event (such as a friend's wedding). (A2+) |
| PW2: Creative Writing | Can write simple phrases and sentences about themselves and imaginary people, where they live and what they do, their family and their hobbies. I can briefly introduce myself in a letter with simple phrases and | Can compose a series of simple phrases and sentences about their family, living conditions, educational background, present or most recent job. Can compose short, simple imaginary biographies and simple poems about people. |

| EBCL Descriptors | Level A1 | Level A2 |
|---|---|---|
| | sentences (family, school, job, hobbies). (A1+) | Can compose about everyday aspects of his environment e.g. people, places, a job or study experience in linked sentences. (A2+) Can compose very short, basic descriptions of events, past activities and personal experiences. (A2+) Can compose a description of a real or imaginary experience (a travelogue) (A2+) |
| IW1: Overall Written Interaction | Can ask for or pass on personal details in written form. I can write a note to tell somebody where I am or where we are to meet. (A1+) | Can compose short, simple formulaic notes relating to matters in areas of immediate need. Can compose in short simple sentences one's wishes, impressions, etc. about such topics as the menu or the taste of the food, on a form to put in a suggestion box found in, for example, a company or university cafeteria. (A2+) |
| IW2: Correspondence | Can write *and answer to* a short simple postcard. I can write *very* short and simple letters or e-mails, telling about everyday things to people I know well. (A1+) | Can compose very simple personal letters expressing thanks and apology. Can compose a simple letter to a friend to introduce myself, my family, school and hobbies.) Can compose short letters or emails, telling about everyday things to people I know well. Can compose a simple personal letter to invite someone or to propose something. (A2+) |
| IW3: Notes, Messages & Forms | Can write numbers and dates, own name, nationality, place of birth/ residence, age, date of birth or arrival in the country etc. such as on a hotel registration form. Can ask for or pass on personal details in written form. Can write a short simple note or message (e.g., to tell somebody where I am or where to meet, to say that someone telephoned, arranging to meet someone, explaining absence). (A1+) | Can compose short, simple notes and messages relating to matters in areas of immediate need. Can fill in a questionnaire giving an account of my educational background, my job, my interests and my specific skills. Can take a short, simple message provided he/she can ask for repetition and reformulation about familial topics. (A2+) |

**Appendix 5**

*Results of the Gain Brought by the Presence of the List in Reducing the Number of Out-of-List Characters for the Corresponding Level for the GPT-4o Model. Green color indicates a gain (positive gain value). The average ratio is the percentage of characters than do not respect the constraint averaged on ten experiments.*

| GPT-4o | | List Not Given | | List Given | | Gain with List |
|---|---|---|---|---|---|---|
| Level | Task | Average Ratio | std | Average Ratio | std | |
| A1 | RW1 | 9,33 % | 1,36 % | 6,88 % | 1,96 % | 2,45 % |
| A1 | RW2 | 12,79 % | 2,23 % | 11,47 % | 2,39 % | 1,32 % |
| A1 | RW3 | 23,38 % | 5,08 % | 10,89 % | 6,72 % | 12,49 % |
| A1 | RW4 | 16,06 % | 3,64 % | 9,06 % | 6,26 % | 7,00 % |
| A1 | RW5 | 25,37 % | 3,00 % | 11,06 % | 8,60 % | 14,31 % |
| A1 | PW1 | 26,81 % | 0,00 % | 6,12 % | 0,00 % | 20,69 % |
| A1 | PW2 | 12,55 % | | 4,17 % | 4,03 % | 8,38 % |
| A1 | IW1 | 11,19 % | | 2,50 % | 1,45 % | 8,69 % |
| A1 | IW2 | 16,06 % | 2,22 % | 5,06 % | 4,37 % | 11,00 % |
| A1 | IW3 | 7,72 % | 2,30 % | 3,33 % | 3,30 % | 4,39 % |
| A1plus | RW1 | 7,60 % | 1,36 % | 4,78 % | 1,96 % | 2,82 % |
| A1plus | RW2 | 12,69 % | 2,23 % | 5,22 % | 2,39 % | 7,47 % |
| A1plus | RW3 | 18,77 % | 5,08 % | 12,79 % | 6,72 % | 5,98 % |
| A1plus | RW4 | 22,74 % | 3,64 % | 13,93 % | 6,26 % | 8,81 % |
| A1plus | RW5 | 24,76 % | 3,00 % | 17,21 % | 8,60 % | 7,55 % |
| A1plus | PW1 | 3,89 % | 0,00 % | 13,87 % | 0,00 % | -9,98 % |
| A1plus | PW2 | 6,47 % | | 0,38 % | 4,03 % | 6,09 % |
| A1plus | IW1 | 15,87 % | | 3,00 % | 1,45 % | 12,87 % |
| A1plus | IW2 | 4,63 % | 2,22 % | 7,39 % | 4,37 % | -2,76 % |
| A1plus | IW3 | 10,43 % | 2,30 % | 4,55 % | 3,30 % | 5,88 % |
| A2 | RW1 | 1,37 % | 1,36 % | 2,02 % | 1,96 % | -0,65 % |
| A2 | RW2 | 2,24 % | 2,23 % | 2,65 % | 2,39 % | -0,41 % |
| A2 | RW3 | 6,15 % | 5,08 % | 6,36 % | 6,72 % | -0,21 % |
| A2 | RW4 | 5,88 % | 3,64 % | 5,89 % | 6,26 % | -0,01 % |
| A2 | RW5 | 7,71 % | 3,00 % | 7,52 % | 8,60 % | 0,19 % |
| A2 | PW1 | 0,00 % | 0,00 % | 0,00 % | 0,00 % | 0,00 % |
| A2 | PW2 | 0,00 % | | 2,33 % | 4,03 % | -2,33 % |
| A2 | IW1 | 0,00 % | | 0,72 % | 1,45 % | -0,72 % |
| A2 | IW2 | 2,44 % | 2,22 % | 3,41 % | 4,37 % | -0,97 % |
| A2 | IW3 | 3,27 % | 2,30 % | 2,39 % | 3,30 % | 0,88 % |

**Appendix 6**

*Results of the Gain Brought in absence of the List in Reducing the Number of Out-of-List Characters for the Corresponding Level for the GPT4o-mini Model. Green color indicates a gain (positive gain value). The average ratio is the percentage of characters than do not respect the constraint averaged on ten experiments.*

| GPT4o-mini | | List Not Given | | List Given | | |
|---|---|---|---|---|---|---|
| Level | Choice | Average Ratio | std | Average Ratio | std | Gain with List |
| A1 | RW1 | 25,80 % | 2,07 % | 25,89 % | 2,60 % | -0,09 % |
| A1 | RW2 | 20,98 % | 1,85 % | 23,12 % | 2,92 % | -2,14 % |
| A1 | RW3 | 21,88 % | 2,98 % | 22,67 % | 2,56 % | -0,79 % |
| A1 | RW4 | 23,05 % | 3,01 % | 29,30 % | 2,49 % | -6,25 % |
| A1 | RW5 | 18,46 % | 1,95 % | 11,54 % | 6,19 % | 6,92 % |
| A1 | PW1 | 26,31 % | 2,30 % | 25,69 % | 2,48 % | 0,62 % |
| A1 | PW2 | 22,64 % | 1,94 % | 19,61 % | 1,22 % | 3,03 % |
| A1 | IW1 | 21,78 % | 2,18 % | 21,28 % | 2,17 % | 0,50 % |
| A1 | IW2 | 22,64 % | 2,42 % | 13,56 % | 2,94 % | 9,08 % |
| A1 | IW3 | 23,56 % | 2,85 % | 26,79 % | 3,24 % | -3,23 % |
| A1plus | RW1 | 22,97 % | 2,07 % | 14,40 % | 2,60 % | 8,57 % |
| A1plus | RW2 | 21,54 % | 1,85 % | 18,62 % | 2,92 % | 2,92 % |
| A1plus | RW3 | 20,13 % | 2,98 % | 18,25 % | 2,56 % | 1,88 % |
| A1plus | RW4 | 20,55 % | 3,01 % | 22,66 % | 2,49 % | -2,11 % |
| A1plus | RW5 | 18,33 % | 1,95 % | 24,12 % | 6,19 % | -5,79 % |
| A1plus | PW1 | 19,47 % | 2,30 % | 20,16 % | 2,48 % | -0,69 % |
| A1plus | PW2 | 15,25 % | 1,94 % | 15,67 % | 1,22 % | -0,42 % |
| A1plus | IW1 | 19,16 % | 2,18 % | 19,27 % | 2,17 % | -0,11 % |
| A1plus | IW2 | 16,51 % | 2,42 % | 17,60 % | 2,94 % | -1,09 % |
| A1plus | IW3 | 18,73 % | 2,85 % | 17,13 % | 3,24 % | 1,60 % |
| A2 | RW1 | 6,18 % | 2,07 % | 5,93 % | 2,60 % | 0,25 % |
| A2 | RW2 | 4,96 % | 1,85 % | 6,36 % | 2,92 % | -1,40 % |
| A2 | RW3 | 8,25 % | 2,98 % | 6,09 % | 2,56 % | 2,16 % |
| A2 | RW4 | 7,20 % | 3,01 % | 8,50 % | 2,49 % | -1,30 % |
| A2 | RW5 | 8,66 % | 1,95 % | 9,49 % | 6,19 % | -0,83 % |
| A2 | PW1 | 5,79 % | 2,30 % | 5,88 % | 2,48 % | -0,09 % |
| A2 | PW2 | 7,81 % | 1,94 % | 7,60 % | 1,22 % | 0,21 % |
| A2 | IW1 | 6,85 % | 2,18 % | 6,41 % | 2,17 % | 0,44 % |
| A2 | IW2 | 5,54 % | 2,42 % | 7,83 % | 2,94 % | -2,29 % |
| A2 | IW3 | 7,48 % | 2,85 % | 6,53 % | 3,24 % | 0,95 % |

**Appendix 7**

*Number of publications on ("chatBot" and "education" and "learning") vs. ("chatGPT" and "education" and "learning") on PubMed        .*

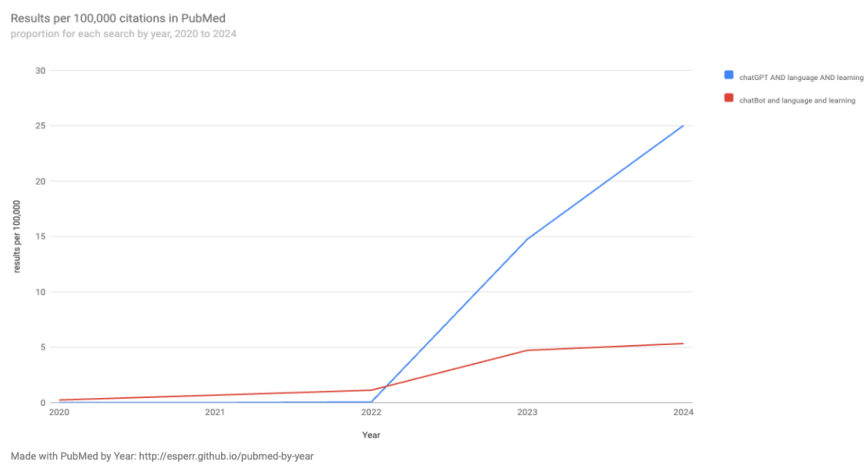

Although this analysis is based on PubMed, a search comparing the number of publications using the terms 'ChatBot AND Education AND Learning' to those mentioning 'ChatGPT AND Education AND Learning' reveals a significant increase in works related to ChatGPT. This trend illustrates a major shift in research on chatbots. The rise of ChatGPT and large language models (LLMs) has redefined the capabilities of chatbots, particularly in the field of education, by enabling more sophisticated, personalized, and contextual interactions. The choice of PubMed stems from the fact that no such analysis tool exists for Google Scholar.

# Authors

*Miao Lin-Zucker* is an Associate Professor at the Linguistics Research Center (CEL) at the University of Lyon 3, specializing in foreign language teaching. Her research focuses on the CEFR applied to Chinese language teaching, Chinese character acquisition and the cognitive processes involved in learning logographic writing systems, with an emphasis on innovative teaching methods.

*Joël Bellassen* is a former university full professor at INALCO and Honorary Inspector General for Chinese at the French Ministry of National Education. His areas of interest include the teaching methods of the Chinese language and writing system, as well as the history of the dissemination of Chinese in France. He is the initiator and co-author of a Dongba script MOOC published online in 2023.

*Jean-Daniel Zucker* is a Senior Researcher in Machine Learning, deputy director of an AI and Mathematical Modeling laboratory and a former Professor of AI at Sorbonne University. His early work focused on using machine learning for Chinese language didactics, and his current research is centered on deep learning and large language models (LLMs).

# 通過提示工程引導 ChatGPT 進行中文作為第二語言的學習：
# 基於 CEFR 和 EBCL 水平的研究

林季苗 里昂第三大學，法國

白樂桑 國立東方語言文化學院，法國

林諸葛 發展研究院，索邦大學，人工智慧和數學建模研究所，法國

**摘要**

自 60 年代以來，聊天機器人在語言學習中的應用有了顯著發展，隨著生成式人工智慧的出現，這些工具已演變為精進的平台。它們現在能夠模擬自然對話，並根據學習者的需求進行課程設計。本研究探討漢語二語學習者如何通過提示工程使用大型語言模型（LLM）作為個人化的聊天機器人，並根據《歐框》（CEFR）和《歐洲漢語基準項目》（EBCL）來規範他們的語言水平。我們重點研究 A1，A1+和 A2 級別，來分析中文教學面臨的獨特挑戰。本研究的目標是開發提示工程，將閱讀和寫作語言技能結合起來，使用高頻字符列表，並控制詞彙的輸出。這些工具由生成式 AI 驅動，旨在通過詞彙和字符的高頻交叉來加強語言實踐。雖然生成式 AI 作為個人化教師展現了潛力，但其效果仍需進一步評估。我們進行了一系列系統性的實驗，使用 ChatGPT 模型來評估它們對提示工程中規定約束條件的遵守程度。結果表明，納入 A1 和 A1+級別的字符，以及相關的參考字符表，顯著提高了對 EBCL 漢字字表的遵守性。在有效的提示下，大型語言模型可以提高學習者的目標語言演練機會，從而幫助增進語言技能。

**關鍵詞**：漢語教學，大型語言模型，提示工程，歐框，歐洲漢語基準項目